\documentclass{article} 
\usepackage{iclr2027_conference,times}

\usepackage{amsmath,amsfonts,bm}

\def\eqref#1{equation~\ref{#1}}

\def\1{\bm{1}}

\DeclareMathAlphabet{\mathsfit}{\encodingdefault}{\sfdefault}{m}{sl}
\SetMathAlphabet{\mathsfit}{bold}{\encodingdefault}{\sfdefault}{bx}{n}

\usepackage{hyperref}
\usepackage{url}
\usepackage{booktabs}
\usepackage{multirow}
\usepackage{enumitem}
\usepackage{caption}

\usepackage{marvosym}

\usepackage{xcolor}
\usepackage[normalem]{ulem}

\definecolor{cvprblue}{rgb}{0.13,0.42,0.85}
\definecolor{linkred}{rgb}{0.85,0.1,0.3}

\hypersetup{
    colorlinks=true,
    citecolor=cvprblue,
    linkcolor=cvprblue,
    urlcolor=linkred,
    filecolor=linkred
}

\title{In-context Robot Learning Made Simple: A Democratized Recipe for Manipulation Tasks}

\newcommand{\authorskipshort}{\hspace{2mm}}

\author{
  \begin{tabular}{l}
  Minxing~Li$^{1*}$ \authorskipshort
  Minghao~Han$^{1*}$ \authorskipshort
  Weizhi~Zhao$^{1}$ \authorskipshort
  Hanwen~Wang$^{1}$ \authorskipshort
  Xiangshuo~Liu$^{1}$\authorskipshort
  \\
  \textbf{Shuyao~Shang}$^{1}$ \authorskipshort
  \textbf{Jingxiang~Zhou}$^{1}$ \authorskipshort
  \textbf{Mingchao~Sun}$^{2}$ \authorskipshort
  \textbf{Hongyu~Pan}$^{2}$ \authorskipshort
  \textbf{Mu~Xu}$^{2}$ \authorskipshort
  \textbf{Yu~Liu}$^{2}\textsuperscript{\dag}$ \authorskipshort
  \\
  \textbf{Lue~Fan}$^{1}\textsuperscript{\dag~\Letter}$ \authorskipshort
  \textbf{Zhaoxiang~Zhang}$^{1}\textsuperscript{\Letter}$
  \end{tabular}
  \\[3mm]
  $^1$NLPR, Institute of Automation, Chinese Academy of Sciences (CASIA) \\
  $^2$Amap, Alibaba Group
  \\[1mm]
{\small\texttt{
\{lue.fan, zhaoxiang.zhang\}@ia.ac.cn
}}
}

\usepackage{wrapfig}
\usepackage{graphicx}
\usepackage{needspace}
\usepackage{multirow}
\usepackage{graphicx}
\usepackage{booktabs}

\iclrfinalcopy 
\begin{document}

\maketitle

\begingroup
\renewcommand\thefootnote{}\footnotetext{$^*$: Equal contribution.\quad \dag: Project leads.\quad \Letter: Corresponding authors.}
\endgroup

\vfill
\begin{figure}[h]
    \centering
    \includegraphics[width=\linewidth]{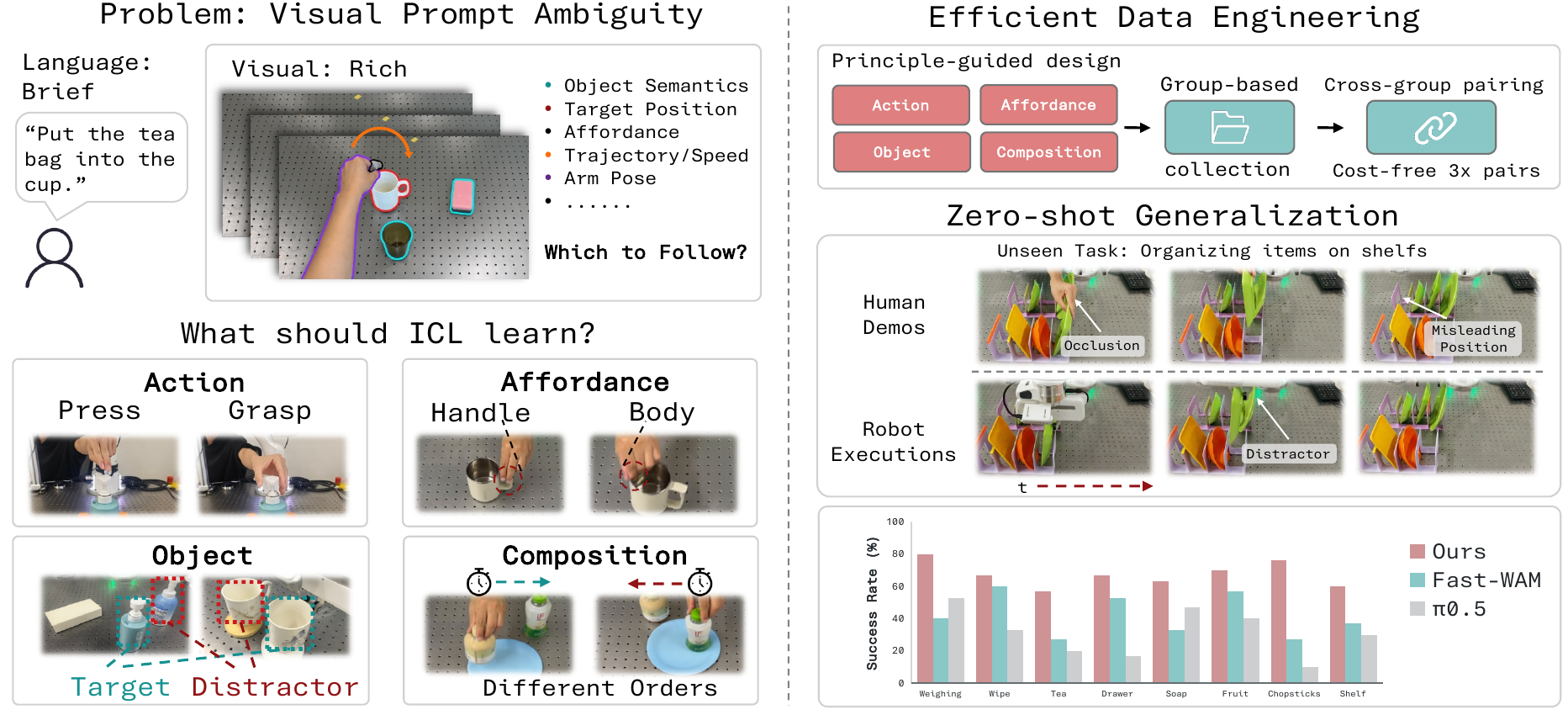}
    \caption{\textbf{Overview of SimpleICL.}
We identify the prompt ambiguity problem in robot ICL and clarify the intrinsic semantics that the model should learn. We further improve task understanding and generalization through efficient data engineering. Our method achieves the best performance across eight evaluation tasks.}
    \label{fig:teaser}
\end{figure}
\vfill

\begin{abstract}
We study robotic in-context learning (ICL), an emerging paradigm that enables robots to infer and execute tasks from visual demonstrations. Despite its growing promise, the problem itself remains under-defined: a visual demonstration simultaneously conveys action trajectories, object semantics, manipulation affordances, spatial relations, and task goals, making it unclear what information the robot is actually expected to follow. In this work, we first provide a clear problem definition of robot ICL that explicitly defines its learning target and resolves this fundamental prompt ambiguity. Building on this definition, we develop a minimalist and reproducible ICL framework (\textbf{SimpleICL}) with a visual prompt encoder and a low-cost data collection protocol. Without massive pre-training or specialized data infrastructure, our framework achieves strong performance in both simulation and real-world environments. Extensive experiments further reveal several key properties of robot ICL, including semantic discrimination and task-relevant disentanglement. We will fully open-source our data and training pipeline to facilitate systematic and reproducible research on robot ICL. The project page can be found at \href{https://simpleicl.github.io/simpleicl/}{\underline{https://simpleicl.github.io/simpleicl/}}.
\end{abstract}

\section{Introduction}\label{sec:intro}

In recent years, the learning paradigm of embodied intelligence has undergone multiple shifts: from imitation learning reliant on large-scale teleoperation data, to real-world RL post-training, and further to the utilization of egocentric data. This surge of interest naturally begs a fundamental question: \emph{what is the most intrinsic way of robot learning?}

Although it is hard to answer such a big question, we can first look back at how humans themselves acquire manipulation skills.
We typically learn a skill by observing others perform it and then imitating them.
Unlike teleoperation-based imitation learning, humans are rarely forced to learn by having someone physically control our arms.
Instead, we infer the task and how it should be performed from visual observations of others. This learning process can be viewed as a form of in-context learning (ICL), where the observations we receive act as visual prompts.
Such visual prompts provide rich task-relevant information—including spatial configurations, object appearance, and fine-grained action trajectories—that is difficult to fully specify through language alone. This rich contextual information makes visual-prompt-based ICL particularly promising for zero-shot adaptation to novel tasks and environments.

In very recent months, several concurrent works have independently begun exploring in-context robot learning, such as HOST~\citep{chen2026robots}, GEN 1.5~\citep{generalist2026gen15}, and Zero-WAM~\citep{zhou2026zero}.
These works have demonstrated preliminary ICL capabilities, achieving zero-shot generalization across novel tasks.
However, as an under-explored learning paradigm, in-context robot learning still harbors numerous open problems.
A central one is prompt ambiguity: video prompts are highly informative, but precisely because they contain so much information, they do not directly specify which aspects of the demonstration the robot should follow and which aspects it should ignore.
For instance, should the robot strictly follow the human's motion trajectory, or should it adapt based on the high-level task/object semantics? Should it strictly replicate the specific way an object is manipulated, or focus solely on achieving the final goal? Figure~\ref{fig:teaser} illustrates this inherent ambiguity in ICL problems. 
Beyond this, several other open questions remain, including but not limited to:
(1) What are the minimal necessary factors for ICL to work in robotics?
(2) How should data for ICL be designed and collected?
(3) Can ICL learn fine-grained manipulation affordances?

To address these questions, we present a simple yet effective framework for robot ICL, termed \textbf{SimpleICL}, adhering to the principles of being simple and easily reproducible.
Within this framework, to tackle the critical challenge of prompt ambiguity, we first establish a problem definition for robot ICL -- what exactly we expect ICL to learn from human demonstrations. 
This definition logically eliminates intent ambiguity during context learning.
The definition naturally induces a corresponding data recipe: the dataset should explicitly separate the factors that define task intent while breaking shortcuts based on irrelevant correlations. Based on this view, we construct a standardized protocol for data design and collection that emphasizes semantic discrimination and robustness to irrelevant variations. We also adopt a simple but effective prompt-conditioned policy architecture to facilitate in-context learning. Overall, our minimalist framework enables us to achieve promising ICL performance at a low data cost, without relying on massive pre-training or specialized data infrastructure.
This holds significant value for democratizing and decentralizing ICL research across the broader community.

In summary, our contributions are threefold:
\begin{itemize}
    \item \textbf{Identifying and addressing prompt ambiguity:} We are the first to expose and thoroughly examine the issue of prompt ambiguity in visual-prompt-based ICL, offering a clear problem definition of ICL to resolve it.
    \item \textbf{A simple, minimalist architecture:} We propose a minimalist robot ICL model architecture without bells and whistles and conduct extensive experiments in both simulation and real-world environments. Our results demonstrate the effectiveness of ICL, while uncovering a range of underlying properties and open challenges for the first time.
    \item \textbf{Low-cost and reproducible data recipe:} We contribute a low-cost, easily reproducible data infrastructure for ICL and will fully open-source the entire data and training pipeline, facilitating the further adoption and advancement of this emerging research paradigm within the robotics community.
\end{itemize}

\section{Related Work}

\paragraph{Vision-Language-Action and World Action Model Policies}
In previous seminal works, embodied studies generally utilize Vision-Language-Action (VLAs) models and World Action Models (WAMs) as foundation models.
The VLAs~\citep{kim2024openvla, black2024pi_0, intelligence2025pi_, bjorck2025gr00t, zitkovich2023rt, liu2024rdt, bu2025agibot, shukor2025smolvla, team2025gemini, galaxea2025, wen2025dexvla} align the text instruction and visual observations into the latent space of a pretrained large vision-language model, which is finetuned to produce the robotic actions.
In parallel, the WAMs~\citep{du2023learninguniversalpoliciestextguided, wu2023unleashing, zhou2024robodreamer, feng2025vidarembodiedvideodiffusion, bharadhwaj2024gen2act, won2025dualstreamdiffusionworldmodelaugmented, cheang2024gr2generativevideolanguageactionmodel, jang2025dreamgenunlockinggeneralizationrobot, zhao2025cotvlavisualchainofthoughtreasoning, cen2025rynnvla, cen2025WorldVLA, zhou2025act2goalworldmodelgeneral, zheng2025flarerobotlearningimplicit, dreamvla25,zhu2025unifiedworldmodelscoupling, liang2025videogenerators, kim2026cosmos, ge2025, pai2025mimicvideo, lingbot-va2026, bi2025motusunifiedlatentaction, ye2026worldactionmodelszeroshot,yuan2026fastwam} are based on video generation models, which consider future video prediction as a prior to infer the actions.
However, most existing policies rely on task-specific finetuning to generalize to novel tasks~\citep{kim2024openvla,black2024pi_0,team2025gemini}.
To reduce the high costs of robotic data collection and model retraining, recent works investigate methodologies to learn from human demonstration videos.
\vspace{-3mm}
\paragraph{Learning from Human Demonstrations}
Unlike robot demonstrations, human manipulation data typically consist only of video, rather than multimodal signals such as states and actions.
WAM-TTT~\citep{feng2026wam} proposes a test-time adaptation that optimizes a lightweight memory to acquire new skills from human video.
ReCAP~\citep{park2026retrieve} appends the new task demonstration to the retrieval pool and retrieves it when executing.
Very recent works propose in-context learning on robotics~\citep{huang2026roboticl}.
Similar to ICL in large language models (LLMs)~\citep{brown2020language}, robot ICL regards the human demonstration videos as part of the prompt along with text instructions.
The video prompts bring richer information such as spatial constraints, intermediate states, and temporal structure.
In addition to directly using the entire video as context~\citep{zhou2026zero,lingbot-va2026}, another approach, HOST~\citep{chen2026robots}, is to temporally align the video in real time and use the corresponding current frame as the conditioning input.
Nevertheless, robot in-context learning remains an under-explored learning paradigm. Given its novelty, our work aims to address the issues of definition, phenomena, and data design in ICL.
\vspace{2pt}
\section{Method}
\vspace{2pt}

\subsection{Problem Definition of Robot ICL}
\label{problem_definition}

As discussed in Section~\ref{sec:intro}, the main difficulty of robot ICL is that a human video prompt contains multiple types of information entangled together.
Some of this information defines the task the robot should follow, while some is incidental variation that the robot should ignore.
To make this distinction explicit, we define robot ICL through two complementary questions: (1) what should be followed from the human demonstration, and (2) what should be ignored.

We first define the task intent that should be followed across the prompt and execution.
We categorize this intent into four semantic dimensions.
(1) \textbf{Action Semantics:} The ICL model must emulate human behavior at the categorical level of actions, including grasping, placing, rotating, pressing, and translating.
(2) \textbf{Object Semantics:} The model must comprehend that it is manipulating a specific object rather than blindly mimicking spatial motion trajectories.
Specifically, if the object's location during robot execution differs from that in the human demonstration, the model should align its execution with the object's actual pose rather than replicating the demonstrator's spatial trajectory.
(3) \textbf{Compositional Semantics}: The ICL model should discern different structural compositions and sequential orderings of actions within human demonstrations. For instance, grasping one object before grasping another conveys a semantics fundamentally distinct from grasping the latter before the former.
(4) \textbf{Affordance Semantics:} When interacting with an object, the ICL model should learn \emph{how} to manipulate it.
For example, it ought to infer from the demonstration whether to grasp the body of a mug or its handle.
While real-world deployments may occasionally demand ICL to condition on additional domain-specific factors, we posit that these four dimensions suffice to support the vast majority of common manipulation tasks without rendering ICL learning overly rigid.

Conversely, we next specify the factors that should be ignored.
These are variations present in the prompt that do not define the task intent and should not be relied upon during execution.
Specifically, we consider three major types of invariance.
(1) \textbf{Spatial Mismatches:} ICL should be robust to changes in the position of the target object between the human demonstration and the robot execution scene.
The model should therefore follow the object semantics rather than copy the absolute spatial trajectory of the human hand.
However, extreme positional disparities are treated as distinct semantics.
(2) \textbf{Human Action Details:} ICL must remain invariant to low-level details of human actions, such as execution speed, arm morphology, hand appearance, specific grasp angles, and personal reaching habits.
Although certain scenarios might intentionally require fine-grained conditioning on these factors (e.g., matching the execution speed to the human demonstration), within the scope of this paper, we temporarily treat these factors as irrelevant noise to establish precise boundaries and clearly dissect the fundamental properties of ICL.
(3) \textbf{Scene Variations:} ICL should also demonstrate strong robustness to visual scene variations, including lighting conditions, background colors, and non-target distractor objects.

In summary, our formulation partitions prompt information into two roles: task intent that should be followed, and irrelevant variation that should be ignored.
This decomposition provides the basis for the data design described next.

\subsection{Data Design and Collection Protocol}
\label{real_world_data}

\begin{figure}[t]
    \centering
    \includegraphics[width=\linewidth]{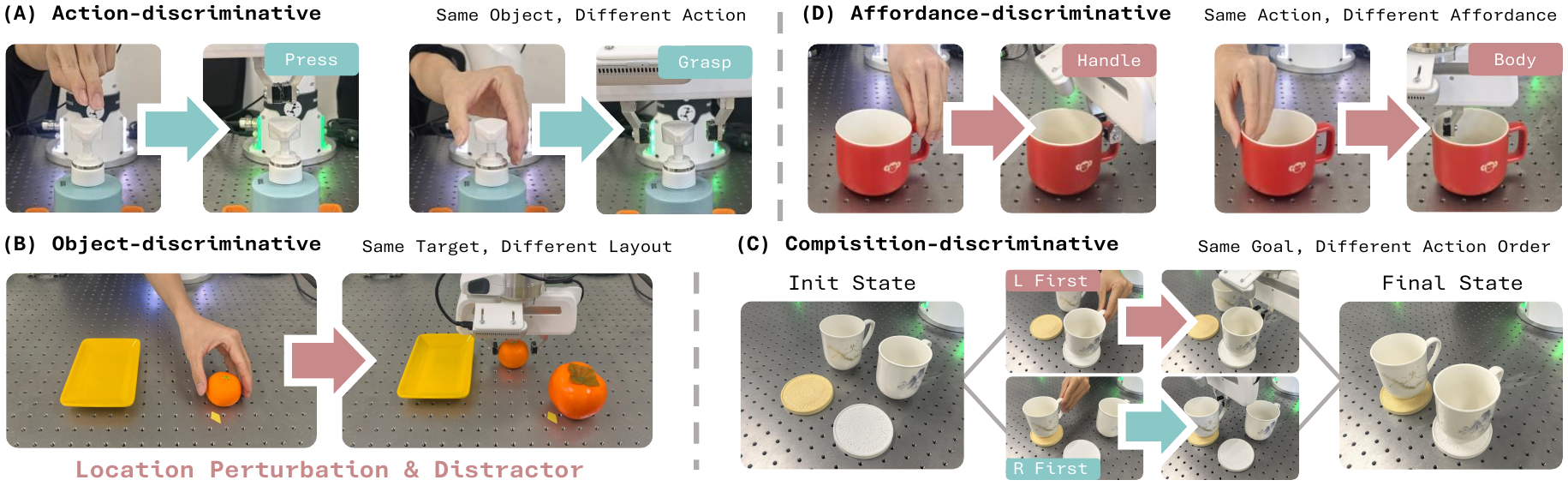}
    \caption{Four types of semantic-discriminative data constructed based on our problem definition.}
    \label{fig:four_data_types}
\end{figure}
The dataset is constructed to reflect the above decomposition.
At the instance level, we create training pairs that force the model to distinguish the semantic factors that define task intent, while breaking correlations that would otherwise let it rely on shortcut cues.
Specifically, we introduce semantic-discriminative data construction (Figure~\ref{fig:four_data_types}) as follows.

\begin{itemize}[leftmargin=*]
    \item \textbf{Action-discriminative data:} To compel the model to distinguish between different action semantics, we collect multiple demonstrations within the same scene using distinct actions. For example, given a bottle of soap dispenser and a tray, one demonstration pair depicts pressing the dispenser, while another depicts grasping the dispenser and placing it on the tray. The robot must differentiate these fine-grained actions and underlying intents.
    
    \item \textbf{Object-discriminative data:} To isolate object identity from scene geometry, we intentionally perturb object positions such that the target object's location during human demonstration differs from that during robot execution. We further introduce semantically distinct distractor objects around the target object. This design encourages the model to identify \emph{which} object should be manipulated rather than follow the human trajectory directly. Additionally, we enforce a strict uniqueness assumption---where only one instance of a specific target object exists per scene---to guarantee semantic unambiguousness.
    
    \item \textbf{Composition-discriminative data:} To capture compositional semantics, we design tasks sharing identical action sets but arranged in different temporal orders (e.g., grasping the banana first then the apple versus grasping the apple first then the banana), thereby compelling the model to attend to sequential dependencies.
    
    \item \textbf{Affordance-discriminative data:} To ground affordance semantics, we establish a human-robot consistency principle within each data pair: the human hand and the robot gripper must maintain identical grasping locations and modes. Across different pairs, however, the manipulation mode varies (e.g., one pair demonstrates grasping the body of a mug, whereas another pair demonstrates grasping its handle).
\end{itemize}

\paragraph{Data augmentation for desired invariances}
To discourage the model from encoding irrelavant factors as task cues, we further apply augmentation along the dimensions that should be ignored.
(1) \emph{Demonstrator invariance.} We collect videos across multiple human demonstrators featuring variations in hand appearance, including bare hands, different colored gloves, and diverse skin tones.
(2) \emph{Environmental invariance.} We systematically vary visual conditions, such as lighting, color temperature, and camera viewpoints in the conditioning videos while pairing them with the same robot execution trajectory, training the model to remain invariant to superficial scene variations.

\paragraph{Group-based collection and augmentation strategy}
The instance-level construction above specifies how task intent and invariance are represented within each pair.
In practice, we implement this through a \emph{collection-group-based strategy}.
Specifically, we define a \textbf{collection group} as a distinct scene coupled with a set of executable tasks within that scene, where each task corresponds to one collection instance.
Upon completing a collection group, we transition to a new group via two types of scene switches: (1) \textbf{Major transitions:} A complete change to an entirely new scene. (2) \textbf{Minor transitions:} Perturbation-based variations within the existing scene, such as altering object placement, adjusting illumination, or swapping background distractor objects.

These minor transitions allow us to pair human demonstrations and robot executions across related but non-identical scene configurations.
As a result, the prompt no longer specifies a one-to-one correspondence that the model can simply copy.
Instead, the policy must recover the task-relevant semantics from the demonstration while remaining robust to spatial perturbations and scene changes.

\begin{figure}[t]
    \centering
    \includegraphics[width=\linewidth]{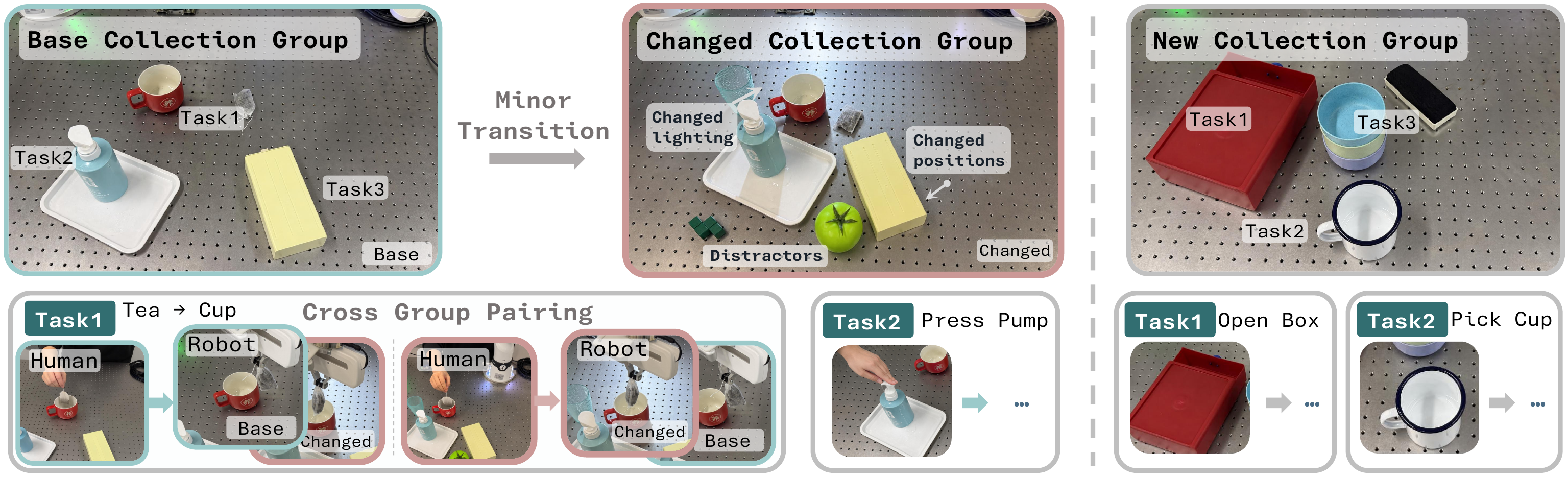}
    \caption{Schematic illustration of the group-based collection and augmentation strategy.}
    \label{fig:data_aug}
    \vspace{-5mm}
\end{figure}

These transitions also allows us to conveniently construct new human-robot data pairs at virtually zero additional collection cost. With our structured design, a single robotic platform can produce up to 3,000 valid human-robot data pairs per working day.
Figure~\ref{fig:data_aug} provides a schematic illustration of the instance construction.

\subsection{Unified Visual Prompt Encoding Module}

Given the above formulation, the role of the prompt encoder is to provide a compact interface between the demonstration video and the policy.
Since the main burden of disambiguation is handled by the problem definition and data construction, the encoder itself can remain simple and generic.

Our encoding module operates as follows.
First, the conditioning video is processed by a pretrained visual encoder to produce a sequence of frame-level embeddings.
These embeddings are projected into a latent space.
We then introduce a small set of learnable \emph{context tokens}, which query the video token sequence through cross-attention and summarize the prompt into a fixed-size representation.
The output of this module is a compact set of conditioned token embeddings.
In practice, as described in Section~\ref{sec:implement}, we concatenate these embeddings into the context of both video and action DiT, which originally contain text and state embeddings.
The context is then fed into the network via cross-attention, resulting in the final policy.

This minimalist instantiation keeps the architecture simple, allowing the empirical study to focus on the effects of prompt semantics and data construction. The simple design also offers several practical advantages.
(1) \emph{Arbitrary-length support}: the number of context tokens is fixed, so the module naturally supports prompts of varying duration.
(2) \emph{Seamless integration}: the resulting representation can be inserted into standard policy backbones with minimal modification.
(3) \emph{End-to-end optimization}: the entire module is jointly optimized with the downstream policy, allowing the prompt representation to adapt to the control objective.

\section{Experiments}

\subsection{Implementation Details.}
\label{sec:implement}
The pretrained Wan2.2-5B~\citep{wan2025} is utilized as the backbone.
Following~\citet{yuan2026fastwam}, we interpolate to obtain a 1B action DiT and set the action horizon to $32$. This interpolated action DiT interacts with the backbone through a Mixture-of-Transformers (MoT) architecture. Images from multiple cameras are concatenated into a single image before being fed into the model.
$128$ query tokens conduct cross-attention via the query-based module on the latent of the conditioning video, which is then fed into the context of both video and action DiTs.
The query-based module stacks $4$ blocks with a hidden dimension of $3072$, resulting in 0.6B parameters.
The video DiT is trained utilizing LoRA, while the other components are trained with full parameters.

\subsection{Simulation Experiments}

\begin{figure}[t]
    \centering
    \includegraphics[width=1\linewidth]{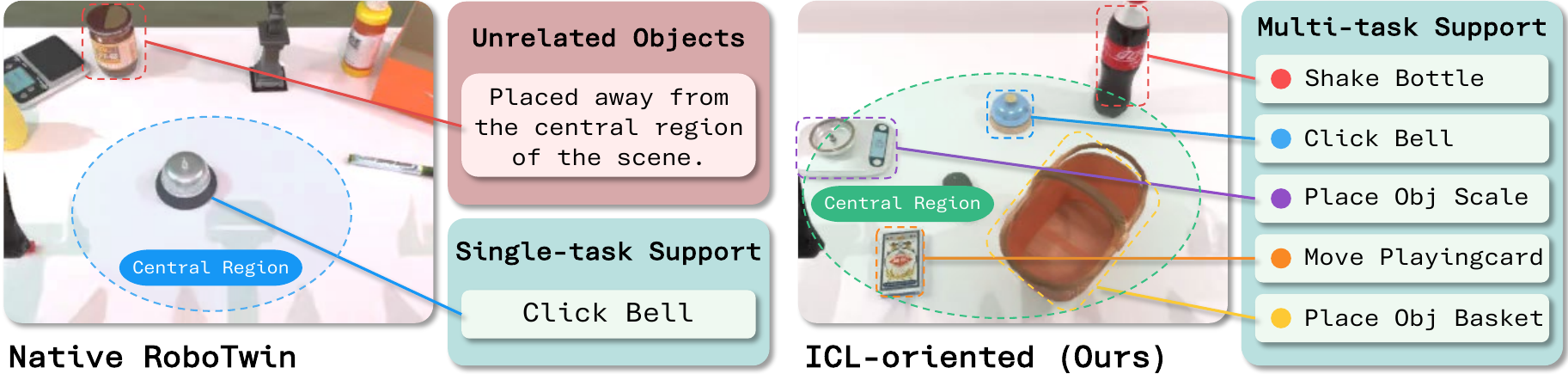}
    \caption{Left: Native RoboTwin data. Right: Proposed ICL-oriented data.}
    \label{fig:sim}
\end{figure}

We first use simulation to validate whether the data principles introduced in Section~\ref{real_world_data} indeed promote semantic discrimination and support robot in-context learning under controlled settings.

\begin{wraptable}{r}{0.5\textwidth}
    \centering
    \vspace{-4mm}
    \caption{Simulation evaluation results on the semantic discrimination test of RoboTwin tasks.}
    \label{tab:sim_native}
    \begin{tabular}{lcc}
        \toprule
        \textbf{Model} &
        \textbf{Trained on} &
        \textbf{Average} \\
        \midrule
        Fast-WAM & Native & 0.38\\
        \midrule
        \multirow{2}{*}{SimpleICL (Ours)}
        & Native & 0.40\\
        & ICL-oriented & \textbf{0.72}\\
        \midrule
        w/o. video cond.
        & ICL-oriented & 0.62\\
        w/o. future pred.
        & ICL-oriented & 0.20\\
        \bottomrule
    \end{tabular}
    \vspace{-2mm}
\end{wraptable}


\paragraph{Training data.}
We instantiate the data construction protocol in Section~\ref{real_world_data} within the RoboTwin 2.0 simulation framework~\citep{chen2025robotwin}. Rather than directly using the native data of RoboTwin, we reconstruct an ICL-oriented training set according to our semantic-discriminative principles. As shown on the left of Figure~\ref{fig:sim}, the native RoboTwin data are primarily designed for standard policy learning, where unrelated objects are often placed outside the central interaction region. While suitable for conventional training, such a setup does not sufficiently require the model to distinguish among multiple task-relevant candidates within the same scene.

Following Section~\ref{real_world_data}, we instead build scenes that simultaneously contain multiple task-relevant objects within the robot's reachable workspace, and collect multiple executable tasks under the same scene with diverse object placements and distractors. This design requires the model to discriminate action and object semantics within a shared visual context, while scene perturbations and cross-group pairing introduce the desired invariances to spatial mismatches and environmental variations. As shown on the right of Figure~\ref{fig:sim}, this yields a simulation benchmark that is better aligned with the intended capability of robot ICL.

Using this protocol, we obtain 18,000 ICL-oriented demonstrations for training. For comparison, we also train our model on 27,500 native RoboTwin demonstrations following prior works~\citep{yuan2026fastwam,bi2025motusunifiedlatentaction,lingbot-va2026}.

\begin{wraptable}{r}{0.5\textwidth}
    \centering    
    \vspace{-4mm}
    \caption{Simulation evaluation results on novel (OOD) tasks. VC. refers to Video Condition. FP. refers to Future Prediction. Tasks are abbreviated.}
    \label{tab:sim_ood}
    \setlength{\tabcolsep}{2.5pt}
    \footnotesize
    \begin{tabular}{lcccc}
        \toprule
        \textbf{Task}& Fast-WAM & Ours& w/o. VC.& w/o. FP.\\
        \midrule
        Cup\_Basket& 0.18& 0.73& 0.46& 0.18\\
        Touch\_Toycar& 0.08& 0.58& 0.08& 0.00\\
        Mouse\_Bowl& 0.26& 0.63& 0.20& 0.17\\
        Block\_Pad& 0.03& 0.50& 0.13& 0.03\\
        Close\_Laptop& 0.33& 0.63& 0.13& 0.23\\
        Toycar\_Box& 0.40& 0.90& 0.37& 0.17\\
        Cards\_Plate& 0.43& 0.63& 0.47& 0.37\\
        \midrule
        \textbf{Average}& 0.25& \textbf{0.66}& 0.26& 0.18\\
        \bottomrule
    \end{tabular}
    \vspace{-3mm}
\end{wraptable}

\paragraph{Evaluation Protocol.}
For evaluation, we utilize our ICL-oriented data pipeline combining multiple RoboTwin tasks to form the semantic discrimination test.
We further design 7 novel (OOD) tasks (Figure~\ref{fig:sim_ood_tasks}) to evaluate our models that are neither natively supported by RoboTwin nor present in the training set.
These tasks for evaluation are produced under ICL-oriented data pipeline as well.
For the sake of the experiment, we utilize \textsc{AgileX} for both the executor entity and the demonstrator (conditioning) platform. 

\paragraph{Results and Findings.}
The quantitative results are reported in Table~\ref{tab:sim_native} and~\ref{tab:sim_ood}. Success rates are reported for each method. Based on the results, we have the following findings.

\begin{itemize}[leftmargin=*]
    \item \textbf{ICL-oriented data improves semantic discrimination in shared scenes.}
    Table~\ref{tab:sim_native} shows that models trained on ICL-oriented data achieve substantially higher success rates on the semantic discrimination test than those trained on native RoboTwin data. This confirms that the data protocol in Section~\ref{real_world_data} better supports the intended discrimination capability of robot ICL.
    
    \item \textbf{Video conditioning enables in-context specification of novel tasks.}
    Although the model without video conditioning performs reasonably well on the semantic discrimination test, its performance drops sharply on novel OOD tasks (Table~\ref{tab:sim_ood}). This indicates that video conditioning is essential for specifying unseen tasks from human demonstrations, rather than merely improving scene-level discrimination.

    \item \textbf{Future prediction is crucial for effective video-conditioned ICL.}
    Removing future prediction causes substantial drops on both the semantic discrimination test and the OOD tasks. This suggests that future prediction is critical for extracting and leveraging task-relevant temporal information from video demonstrations.
\end{itemize}

\subsection{Real-World Experiments}

We instantiate the real-world data collection and evaluation pipeline following the framework established in Section~\ref{real_world_data}. We conduct experiments using \textsc{Franka} robotic platform, equipped with three Intel RealSense D435 cameras. The teleoperation is conducted with 3D connexion space mouse.

\subsubsection{Main Results}

\begin{table*}[t]
    \centering
    \caption{Real-world evaluation on eight unseen tasks.  Success rates are reported for each method. Task abbreviations:  \textbf{Wipe} = wiping a plate; \textbf{Tea} = serving tea; \textbf{Drawer} = pulling a drawer; \textbf{Soap} = pressing a soap dispenser; \textbf{Fruit} = placing fruit; \textbf{Chopsticks} = organizing chopsticks; \textbf{Shelf} = organizing items on a shelf. Please refer to appendices for detailed task design.}
    \label{tab:real_results}
    \renewcommand{\arraystretch}{1.2}
    \resizebox{.99\textwidth}{!}{
\begin{tabular}{clccccccccc}
    \toprule
    & \textbf{Method} & Weighing & Wipe & Tea & Drawer & Soap & Fruit & Chopsticks & Shelf & Average \\
    \midrule

    \multirow{3}{*}{\rotatebox[origin=c]{90}{\textbf{Easy}}}
    & \textsc{Fast-WAM} 
    & 0.90 & 0.87 & 0.83 & 0.80 & 0.63 & 0.90 & 0.73 & 0.77 & 0.80 \\

    & $\pi_{0.5}$ 
    & 0.87 & 0.97 & 0.80 & 0.60 & 0.73 & 0.60 & 0.53 & 0.67 & 0.72 \\

    & \textbf{SimpleICL (ours)} 
    & 0.93 & 0.77 & 0.73 & 0.83 & 0.73 & 0.87 & 0.80 & 0.83 & \textbf{0.81} \\

    \midrule

    \multirow{3}{*}{\rotatebox[origin=c]{90}{\textbf{Medium}}}
    & \textsc{Fast-WAM} 
    & 0.83 & 0.50 & 0.63 & 0.43 & 0.60 & 0.83 & 0.63 & 0.73 & 0.65 \\

    & $\pi_{0.5}$ 
    & 0.63 & 0.47 & 0.37 & 0.43 & 0.57 & 0.47 & 0.23 & 0.57 & 0.47 \\

    & \textbf{SimpleICL (ours)} 
    & 0.83 & 0.67 & 0.60 & 0.80 & 0.67 & 0.63 & 0.73 & 0.63 & \textbf{0.70} \\

    \midrule

    \multirow{3}{*}{\rotatebox[origin=c]{90}{\textbf{Hard}}}
    & \textsc{Fast-WAM} 
    & 0.40 & 0.60 & 0.27 & 0.53 & 0.33 & 0.57 & 0.27 & 0.37 & 0.42 \\

    & $\pi_{0.5}$ 
    & 0.53 & 0.33 & 0.20 & 0.17 & 0.47 & 0.40 & 0.10 & 0.30 & 0.31 \\

    & \textbf{SimpleICL (ours)} 
    & 0.80 & 0.67 & 0.57 & 0.67 & 0.63 & 0.70 & 0.77 & 0.60 & \textbf{0.68} \\

    \bottomrule
    \vspace{-8mm}
\end{tabular}
}
\end{table*}

\begin{figure}[t]
    \centering
    \includegraphics[width=\linewidth]{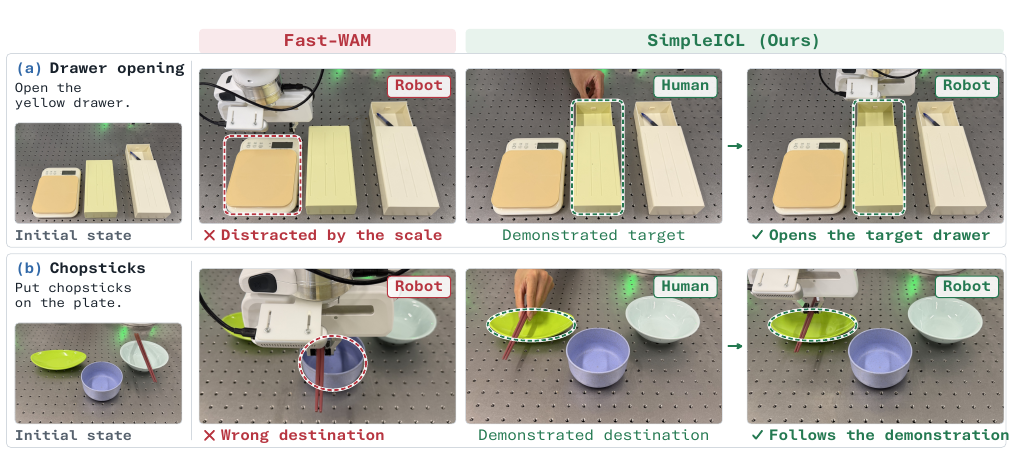}
    \caption{Qualitative comparison between our method and existing baselines. At the \textbf{Medium} level (top), \textsc{Fast-WAM} is distracted by a nearby scale, while our method correctly locates the drawer. At the \textbf{Hard} level (bottom), \textsc{Fast-WAM} mistakenly places chopsticks onto the round bowl, whereas ours reaches the correct destination.}
    \label{fig:qualitative}
\end{figure}

\begin{figure}[t]
    \centering
    \includegraphics[width=\linewidth]{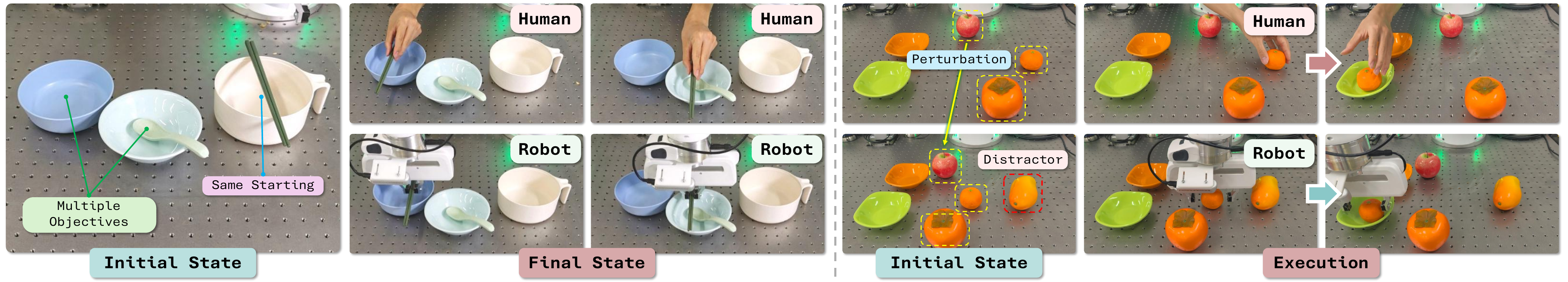}
    \caption{Qualitative examples showing that the ICL model follows the semantic identity of the target objects. Left: the robot matches the human-selected bowl. Right: the robot still succeeds even after the target fruit is displaced and a similar distractor takes over its original position.
}
    \label{fig:qualitative_obj}
\end{figure}

\paragraph{Real-world Task Setup.}
We evaluate our method on eight unseen novel tasks. These tasks are: \emph{weighing, wiping a plate, serving tea, pulling a drawer, pressing a soap dispenser, placing fruit, organizing chopsticks, and organizing items on a shelf}. Critically, the sponge and plate used in the wiping task, the drawer in the pulling task, the fruit and bowls in the placing task, the items to be organized in the shelf task, the soap dispenser being pressed, and the target teacup in the tea-serving task are all objects that never appeared in the training set. This ensures that the evaluation genuinely tests the model's ability to generalize to novel objects and tasks through in-context learning from human demonstrations.
\vspace{-3mm}
\paragraph{Evaluation Protocol.}
We measure model capability across three levels of difficulty. \textbf{Easy}: the scene contains only one executable task with a unique way to perform it. This is the most commonly used evaluation setting in current literature. \textbf{Medium}: the scene contains multiple distinct executable tasks, requiring the model to possess discriminative capability to select the correct task. \textbf{Hard}: the task itself admits multiple valid execution modes. For example, in the plate-wiping task, two plates are placed adjacent to each other, and the model must correctly determine which one to wipe. This level demands the strongest discrimination ability and is the most challenging.
\vspace{-3mm}
\paragraph{Results and Findings.}
The quantitative results are reported in Table~\ref{tab:real_results}, organized into these three levels. Each entry represents the success rate of the corresponding method. We compare our method against two baselines: \textsc{Fast-WAM} and $\pi_{0.5}$. Figure~\ref{fig:qualitative} presents a qualitative comparison between our method and existing baselines. Based on these results, we have the following findings.
\begin{itemize}[leftmargin=*]
    \item \textbf{ICL model has better generalization to novel objects and scenes.} At the Easy Level, where the scene contains only a fixed task, the baseline methods also perform reasonably well, but our model achieves a higher success rate. This indicates that visual in-context learning endows the model with better generalization.
    
    \item \textbf{ICL model better disentangles scene observations from conditioned task intents}. In the medium-level and hard-level tasks, the same observation may correspond to multiple task intents. Language-conditioned models, such as Fast-WAM, often exhibit shortcut learning, binding a particular action intent to a fixed scene pattern rather than properly interpreting the conditioning signal. Figure~\ref{fig:qualitative} illustrates qualitative examples.

    \item \textbf{ICL model can accurately identify the semantic identity of the target object}, rather than simply imitating actions or memorizing spatial locations. In the hard-level tasks, we introduce distractor objects that are highly similar to the target in both spatial location and semantic identity. Even under such challenging conditions, our ICL model maintains a high task success rate. Figure~\ref{fig:qualitative_obj} illustrates qualitative examples.
\end{itemize}

\subsubsection{Analysis of Semantic Learning}
In this section, we further conduct a dedicated analysis of \textbf{SimpleICL}'s semantic learning.

\paragraph{Experiment Control of Action.}
To analyze action semantics, we collect data where the same object is manipulated with different actions. As a control, for each object, we keep only one action in training, such that object identity is correlated with a single action. For example, the red chopsticks are always rotated, while the green chopsticks are always placed onto another bowl.
For evaluation, we consider three representative objects, each supporting multiple action modes: chopsticks (rotated by \(90^\circ\) or placed onto another bowl), a tissue (pressed or pushed forward), and a sponge (pressed or moved to wipe a plate).

\paragraph{Experiment Control of Composition.}
For the analysis of composition, we collect training data where tasks are executed in the same scene in different temporal orders. For example, we place two visually distinct apples into a fruit basket in different orders, and the two demonstrations with different manipulation orders are all recorded as training data.
Conversely, for comparison, we disable composition discriminative data by only incorporating a specific order of demonstration in a scene into the training set.
For evaluation, we evaluate on the task of sequentially organizing items on a table, using novel items never seen during training: snacks, cups, and bags. 

\paragraph{Experiment Control of Affordance.}

For the analysis of affordance, we collect data where the same object is manipulated multiple times with different affordances. To disable this strategy, for each object, we only collect demonstrations involving the same affordance for training. 
For evaluation, we use three classic objects, each supporting multiple affordances: a tape measure (grasping the ring or the body), a cup with a handle (grasping the handle or the body), and a roll of tape (grasping the left side or the right side). 

\paragraph{Findings.}
Table~\ref{tab:composition_affordance} shows the experimental results, where we have the following findings.

\begin{table*}[t]
    \centering
    \caption{Analysis of action, composition, and affordance under discriminative data collection (ours) versus without discriminative data collection. The metric is the intent-following rate. \dag: DC denotes discriminative collection.}
    \label{tab:composition_affordance}
    \renewcommand{\arraystretch}{1.0}
    \resizebox{\textwidth}{!}{
    \begin{tabular}{lcccccccccc}
        \toprule
        \multirow{2}{*}{\textbf{Setting}}
        & \multicolumn{3}{c}{\textbf{Action}}
        & \multicolumn{3}{c}{\textbf{Composition}}
        & \multicolumn{3}{c}{\textbf{Affordance}}
        & \multirow{2}{*}{\textbf{Average}} \\
        \cmidrule(lr){2-4} \cmidrule(lr){5-7} \cmidrule(lr){8-10}
        & Chopsticks & Tissue & Sponge
        & Snack & Cup & Bag
        & Cup & Measure & Tape & \\
        \midrule
        w/. DC\dag
        & 0.93 & \textbf{0.83} & \textbf{0.90}
        & \textbf{0.97} & \textbf{1.00} & 0.90
        & \textbf{0.87} & \textbf{0.77} & \textbf{0.83}
        & \textbf{0.89} \\
        w/o. DC
        & \textbf{0.97} & 0.67 & 0.83
        & 0.93 & 0.97 & \textbf{0.93}
        & 0.60 & 0.40 & 0.47
        & 0.75 \\
        \bottomrule
    \end{tabular}
    }
\end{table*}

\begin{itemize}[leftmargin=*]
\item Our model demonstrates strong \textbf{semantic discrimination} capabilities, allowing it to effectively distinguish and follow different action, composition and affordance patterns exhibited in human demonstrations.

\item For \textbf{composition discrimination}, it is unnecessary to explicitly construct data with different compositions for the same scene. The ICL model can spontaneously learn to disentangle scene content from composition patterns directly from the training data.

\item In contrast, for \textbf{affordance discrimination}, training data with different affordances must be explicitly constructed for the same object to force the model to acquire this capability. Under our current setting, the ICL model cannot spontaneously disentangle object identity from affordance. This indicates that within the scope of ICL, fine-grained affordance may be harder to learn than global temporal and compositional relationships.

\item For \textbf{action discrimination}, the model retains moderate discrimination ability even without discriminative data collection, indicating partial spontaneous disentanglement of object identity and action type. However, the remaining confusion suggests that this disentanglement is incomplete. This indicates that action learning may lies between composition and affordance in difficulty within robot ICL.

\end{itemize}
\begin{wrapfigure}{r}{0.60\textwidth}
    \vspace{-5mm}
    \centering
    \captionof{table}{Robustness analysis. Success rates are reported under multiple disturbance categories. \textbf{Spatial} denotes that the target object position during robot execution is shifted relative to that in the human demonstration. \textbf{Human} denotes human action details, where \textbf{App.} changes the appearance of the demonstrator's hand (e.g., wearing gloves), while \textbf{Sty.} changes the human subject and motion style. For \textbf{Scene Variations}, \textbf{Light} changes the illumination condition, \textbf{Bg.} introduces background color distractions, and \textbf{Clut.} adds irrelevant clutter objects to the scene. \dag: CP means cross-group pairing strategy for data augmentation.}
    \label{tab:disturbance}
    \resizebox{\linewidth}{!}{
    \begin{tabular}{lcccccccc}
        \toprule
        \multirow{2}{*}{\textbf{Method}} 
        & \multirow{2}{*}{\textbf{Orig.}} 
        & \multirow{2}{*}{\textbf{Spatial}} 
        & \multicolumn{2}{c}{\textbf{Human}} 
        & \multicolumn{3}{c}{\textbf{Scene Variations}} 
        & \multirow{2}{*}{\textbf{Avg.}} \\
        \cmidrule(lr){4-5}
        \cmidrule(lr){6-8}
        & & & App. & Sty. & Light & Bg. & Clut. & \\
        \midrule
        \textbf{Ours}
        & \textbf{0.81}
        & 0.79
        & 0.77 & 0.80
        & 0.78 & 0.79 & 0.75
        & \textbf{0.78} \\
        w/o. CP\dag
        & 0.80
        & 0.51
        & 0.78 & 0.79
        & 0.73 & 0.71 & 0.66
        & 0.70 \\
        \bottomrule
    \end{tabular}
    }
    \par\vspace{6mm}
    \includegraphics[width=\linewidth]{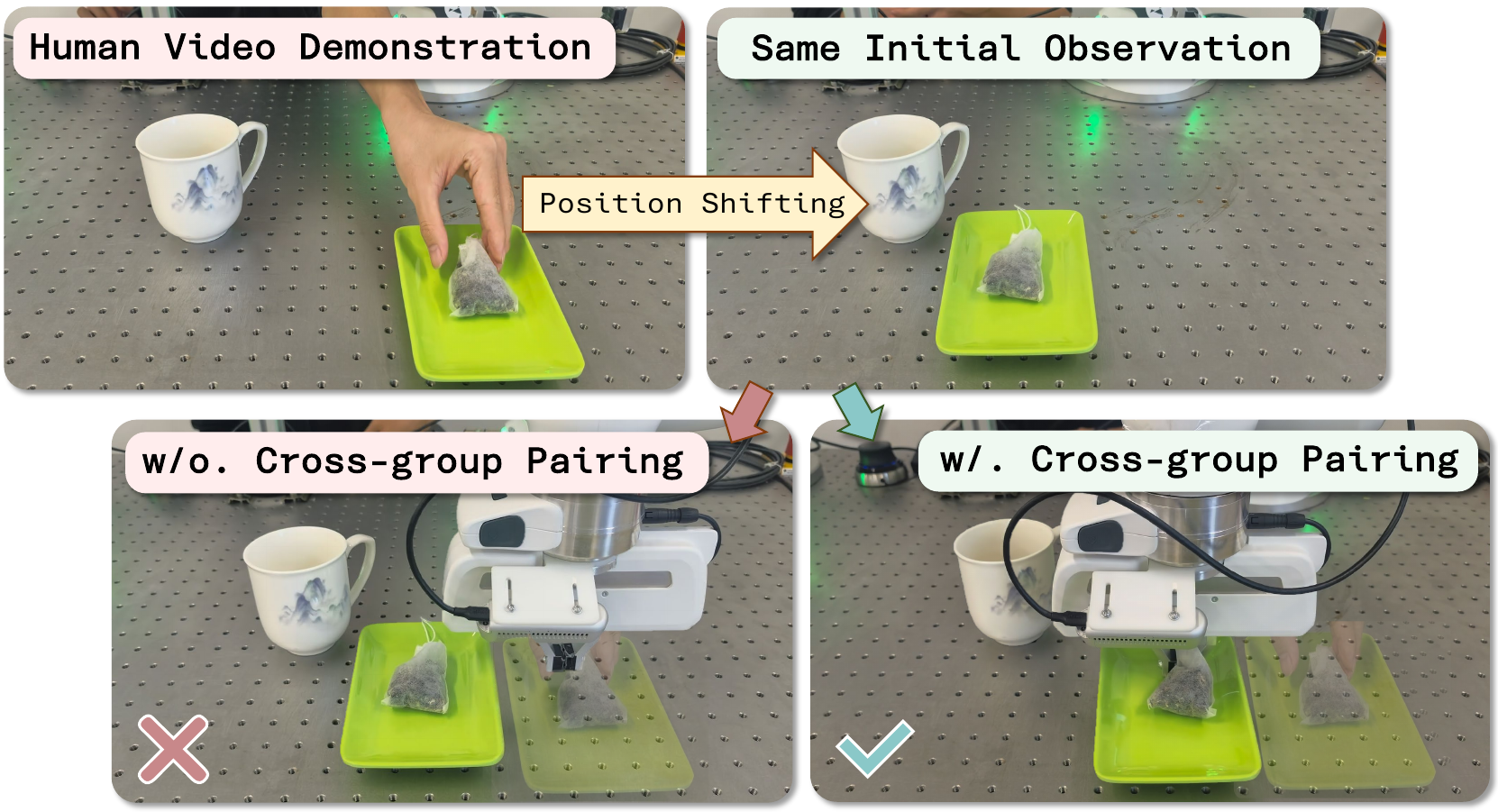}
    \captionof{figure}{The effect of our grouped pairing strategy. Our model correctly identifies the target object despite its position shifting relative to the human demonstration. Without grouped pairing, the model naively follows the position of the object shown in the context.}
    \label{fig:ablation}
    \vspace{-5mm}
\end{wrapfigure}

\subsubsection{Robustness Analysis}

As defined in Section~\ref{problem_definition}, robot ICL should not only capture task-relevant semantics, but also remain robust to three categories of task-irrelevant variations: \textbf{spatial mismatches}, \textbf{human action details}, and \textbf{scene variations}. We therefore evaluate our model under controlled disturbances corresponding to each category.

As shown in Table~\ref{tab:disturbance}, our model remains robust across all three categories of disturbances, which is consistent with the invariance requirements in our problem definition. In particular, the model maintains strong performance under changes in demonstrator appearance/style and scene-level visual variations, indicating that it does not over-rely on demonstrator-specific details or superficial environmental cues.

We further analyze the role of the cross-group pairing strategy (CP). Without CP, each robot trajectory is paired with a human video collected under nearly identical object placement and environmental conditions. This encourages the model to spuriously bind task intent to fixed scene layouts and demonstrated spatial configurations. As a result, performance drops significantly under \textbf{Spatial}, and also decreases under \textbf{Bg.} and \textbf{Clut.}. These results indicate that the proposed cross-group pairing is important not merely for adding data diversity, but for explicitly enforcing the desired invariances of robot ICL.

Figure~\ref{fig:ablation} provides a qualitative example. When the target object shifts relative to the human demonstration, our full model still grounds its action on the object in the current execution scene. In contrast, without cross-group pairing, the policy mechanically moves toward the position implied by the human video, exhibiting a trajectory-copying failure mode rather than object-aware grounding.

\subsection{Attention Visualization}
\label{attn_viz}

\begin{figure}
    \centering
    \includegraphics[width=1\linewidth]{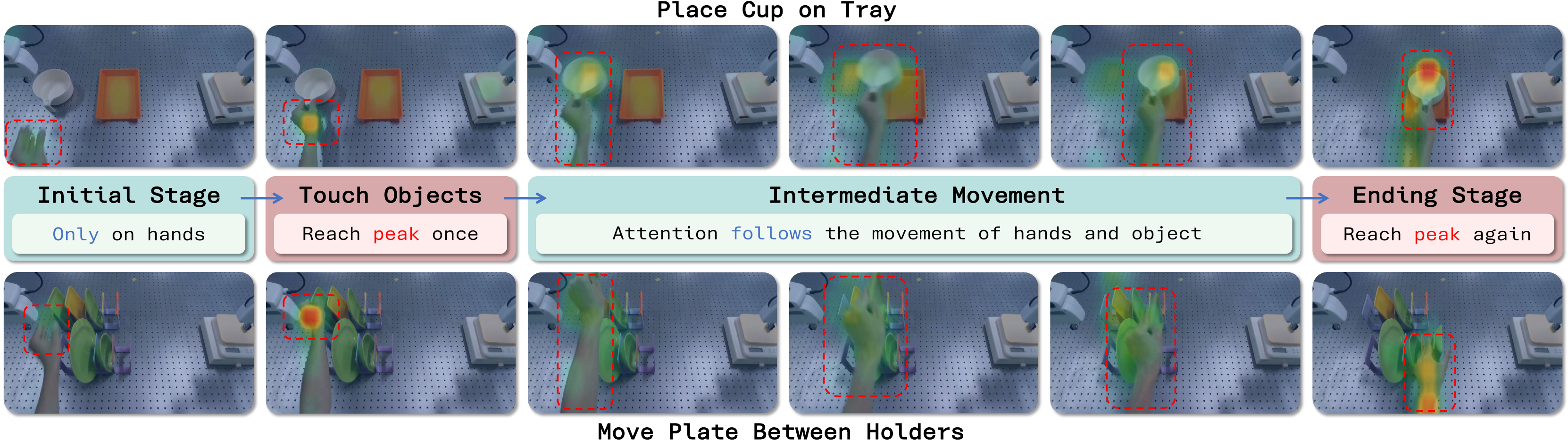}
    \caption{The timeline of attention visualization heat map for the human conditioning videos.}
    \label{fig:attn_viz}
    \vspace{-5mm}
\end{figure}

To better understand whether the policy indeed extracts task-relevant cues, we visualize its attention over the conditioning video.
Figure~\ref{fig:attn_viz} demonstrates a timeline of attention heat map for the human conditioning videos during task execution.
We discover that: 
(a) At the initial stage, the attention only appears on the human hands; 
(b) The peaks of attention intensity occur twice: when the hand first touches on the target object and when the task is completed, respectively; 
(c) During the intermediate stage, the attention follows the movement of both hands and the object with moderate intensity.
These observations suggest that the model exhibits temporally structured and interaction-centric attention.
Rather than uniformly attending to the scene, it focuses on the action-relevant cues at different execution stages, with pronounced responses at key interaction events such as object contact and task completion.

\section{Conclusion}
In this paper, we formulate a series of open questions surrounding the emerging area of robot in-context learning.
We identify prompt ambiguity as a central problem, and establish an explicit definition of robot ICL to resolve this. 
Based on this formulation, we propose \textbf{SimpleICL}, a minimalist and reproducible framework including a specially designed data recipe and a generic module for video conditioning.
Experiments in simulation and the real world validate its effectiveness, offering a simple, low-cost foundation for advancing in-context robot learning.
More importantly, our experiments reveal several insightful findings regarding the nature of robot ICL.
We believe that this field remains a promising and important direction that warrants further investigation.

\newpage
\bibliography{iclr2027_conference}
\bibliographystyle{iclr2027_conference}

\appendix
\newpage
\section{Simulation Novel (OOD) Tasks}

We design 7 novel (OOD) tasks for simulation evaluation; from left to right, they are \textit{Put Toycar in Plasticbox}, \textit{Put Cup in Basket}, \textit{Touch Toycar}, \textit{Place Playingcards Plate}, \textit{Close Laptop}, \textit{Move Block on Pad}, and \textit{Place Mouse in Bowl}.

\begin{figure}[h]
    \centering
    \includegraphics[width=1\linewidth]{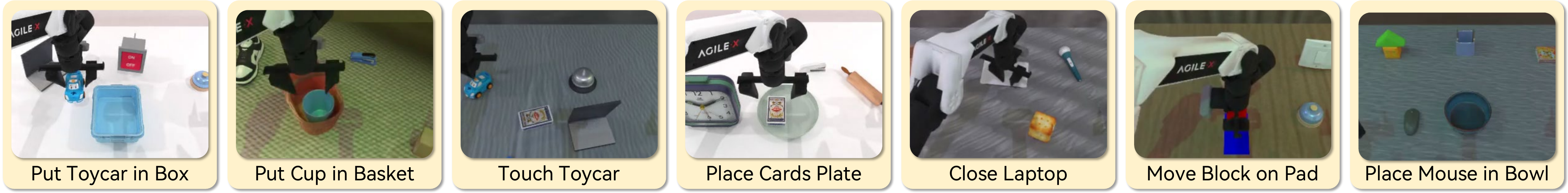}
    \caption{Novel (OOD) tasks for simulation evaluation.}
    \label{fig:sim_ood_tasks}
\end{figure}

\section{Real-world Samples}
\begingroup
\newcommand{\demoframe}[2]{%
  \includegraphics[width=0.196\linewidth]{figs/real_demos/#1-#2.jpg}%
}
\newcommand{\demorow}[1]{%
  \demoframe{#1}{1}\hfill
  \demoframe{#1}{2}\hfill
  \demoframe{#1}{3}\hfill
  \demoframe{#1}{4}\hfill
  \demoframe{#1}{5}\par
}
\newcommand{\demotask}[3]{%
  \par\addvspace{6pt}\noindent
  \begin{minipage}{\linewidth}
    \captionsetup{type=figure,skip=3pt}
    {\footnotesize\textbf{Human demonstration}\hfill Earlier $\longrightarrow$ Later\par}
    \vspace{1pt}
    \demorow{#1-h}
    \vspace{2pt}
    {\footnotesize\textbf{Robot execution}\par}
    \vspace{1pt}
    \demorow{#1}
    \caption{\textbf{#2.}}
    \label{fig:demo_#3}
  \end{minipage}\par
}

\demotask{Weigh.}{Weighing}{weighing}
\demotask{Wipe}{Wiping a plate}{wiping_plate}
\demotask{Tea}{Serving tea}{serving_tea}
\demotask{Drawer}{Pulling a drawer}{pulling_drawer}

\demotask{Soap}{Pressing a soap dispenser}{pressing_soap_dispenser}
\demotask{Fruit}{Placing fruit}{placing_fruit}
\demotask{Chop.}{Organizing chopsticks and bowls}{organizing_chopsticks_bowls}
\demotask{Shelf}{Organizing items on a shelf}{organizing_shelf}
\endgroup

\end{document}